\documentclass[runningheads]{llncs}
\usepackage{graphicx}
\usepackage{xcolor}
\usepackage{array}
\usepackage{caption}
\usepackage{subcaption}
\usepackage{amsmath}
\usepackage{amsfonts}
\usepackage{amssymb}

\newcommand\scalemath[2]{\scalebox{#1}{\mbox{\ensuremath{\displaystyle #2}}}}
\begin{document}
\title{Joint Span Segmentation and Rhetorical Role Labeling with Data Augmentation for Legal Documents}
\titlerunning{Rhetorical Role Labeling for Legal Documents}

\author{Santosh T.Y.S.S \and
Philipp Bock \and
Matthias Grabmair}
\authorrunning{Santosh et al.}
\institute{School of Computation, Information, and Technology; \\ Technical University of Munich, Germany\\
\email{\{santosh.tokala, philipp.bock, matthias.grabmair\}@tum.de}}

\maketitle      
% typeset the header of the contribution
%
\begin{abstract}
Segmentation and Rhetorical Role Labeling of legal judgements play a crucial role in retrieval and adjacent tasks, including case summarization, semantic search, argument mining etc. Previous approaches have formulated this task either as independent classification or sequence labeling of sentences. In this work, we reformulate the task at span level as identifying spans of multiple consecutive sentences that share the same rhetorical role label to be assigned via classification. We employ semi-Markov Conditional Random Fields (CRF) to jointly learn span segmentation and span label assignment. We further explore three data augmentation strategies to mitigate the data scarcity in the specialized domain of law where individual documents tend to be very long and annotation cost is high. Our experiments demonstrate improvement of span-level prediction metrics with a semi-Markov CRF model over a CRF baseline. This benefit is contingent on the presence of multi sentence spans in the document.

\end{abstract}

\section{Introduction}
Rhetorical Role Labeling (RRL) of legal documents involves segmenting a document into semantically coherent chunks and assigning a label to the chunk that reflects its function in the legal discourse (e.g., preamble, fact, evidence, reasoning). RRL for long legal case documents is a precursor task to several downstream tasks, such as case summarization \cite{hachey2006extractive,saravanan2008automatic,kalamkar2022corpus,farzindar2004letsum} , fact-based semantic case search \cite{nejadgholi2017semi}, argument mining \cite{walker2019automatic} and judgement prediction \cite{kalamkar2022corpus}. 

Prior works in RRL on legal judgements have regarded the task either as straightforward classification of sentences without modeling any contextual dependency between them \cite{ahmad2020understanding,walker2019automatic} or as sequence labeling \cite{yamada2019neural,bhattacharya2021deeprhole,ghosh2019identification,kalamkar2022corpus}. Initial works \cite{saravanan2008automatic,farzindar2004letsum,hachey2006extractive} performed RRL using hand-crafted features as part of a summarization pipeline. Savelka et al. \cite{savelka2018segmenting} employed a CRF on hand-crafted features to segment US court decisions into functional and issue specific parts. Similarly, Walker et al. \cite{walker2019automatic} used engineered features  for RRL on US Board of Veterans’ Appeals (BVA) decisions. With the rise of deep learning, Yamada et al.  \cite{yamada2019neural} , Ghosh et al.  \cite{ghosh2019identification}, Paheli et al. \cite{bhattacharya2021deeprhole} and Ahmad et al. \cite{ahmad2020understanding} employed deep learning based BiLSTM-CRF models for RRL on Japanese civil rights judgements, Indian Supreme Court opinions, UK supreme court judgements and the US BVA corpus respectively. More recently, Kalamkar et al. \cite{kalamkar2022corpus} benchmarked RRL on Indian legal documents using a Hierarchical Sequential Labeling Network model (HSLN). The corpus they used claims to be the largest available corpus of legal documents annotated with rhetorical sentence roles. 

In this work we approach RRL on legal documents with the observation that the texts of judgement are not only very long, but also often contain large sections of the same sentence type (e.g. explanations of case facts). We hence build models that  segment the document into thematically coherent sets of contiguous sequence of sentences (which we refer to as \textit{spans}) and assign them labels. %\footnote{In this work, span refers to contiguous sequence of sentences.} 
We also hypothesize that modeling documents at this span level can also help to capture certain types of contexts effectively that may be spread across long sequences of sentences that can be collapsed into a much smaller number of thematically coherent spans. For example, when case documents are to be retrieved according to certain types of information, then aggregating that content from a small number of topical blocks across a long document is intuitive. At the same type, we explore how this assumption of topical continuity in the law can help RRL models learn better from small amounts of training data.

To tackle this problem as sequential span classification, we apply semi-Markov Conditional Random Field (CRF) \cite{sarawagi2004semi}, which have been proposed to jointly handle span segmentation and labeling. %The model takes a sequence of sentences as input and employs a hierarchical BiLSTM to obtain contextual representations for every possible span. A semi-markov CRF layer uses these to estimate the probability for each possible span and its possible labels. The best optimal span sequence along with their rhetorical roles is then determined using the Viterbi algorithm \cite{forney1973viterbi}.  
Semi-Markov CRFs have been used in various tasks such as Chinese word segmentation \cite{liu2016exploring,kong2015segmental}, named entity recognition \cite{ye2018hybrid,zhuo2016segment,arora2019semi}, character-level parts of speech labelling \cite{kemos2019neural}, phone recognition \cite{lu2016segmental}, chord recognition \cite{masada2017chord}, biomedical abstract segmentation \cite{yamada2020sequential} and piano transcription  \cite{yan2021skipping}. Most previous works dealt with shorter input sequences and thus contained smaller span lengths, which allows for a convenient upper bound on the maximum length of a span. In this work, we assess the performance of semi-Markov CRFs on legal judgements, which are usually very long and also possess a potentially large range of labels,  making this setup even more challenging. 

Obtaining sufficiently large amounts of annotated data for deep learning models in specialized domains like the law is very expensive as it requires expert annotators. To mitigate this data scarcity, we explore three strategies of data augmentation (DA) such as random deletion of words, back translation and swapping of sentences within a span. DA techniques which are common in computer vision field, has witnessed growing interest in NLP tasks due to the twin challenge of large annotated data for neural networks and expensive data annotation in low-resource domains \cite{feng2021survey}. %explore data augmentations techniques by automatically creating modified versions of existing annotated data without affecting the rhetorical role of the non-modified sentences. Data augmentation techniques are common  in computer vision (e.g., rotation, mirroring), but less straightforward in text. Recently, it has seen wide interest in NLP tasks due to the twin challenge of  large annotated data for neural networks and expensive data annotation in low-resource domains \cite{feng2021survey}. We explore three strategies of data augmentation such as random deletion of words, back translation, and swapping of sentences within a span.  
In sum, this paper contributes the casting RRL of legal judgments as a sequential span classification task and associated experiments with semi-Markov CRFs on existing public datasets. We also explore three data augmentation strategies to assess their impact on the task. Our experiments demonstrate that our semi-Markov CRF model performs better compared to a  CRF baseline on documents characterized by multi-sentence spans. \footnote{Our code is available at https://github.com/TUMLegalTech/Span-RRL-ECIR23}

\section{Method}
Our hierarchical semi-Markov CRF model takes the judgement document $x = \{x_1, x_2, \ldots, x_m\}$ as input, where $x_i = \{x_{i1}, x_{i2},\ldots,x_{in}\}$ and outputs the rhetorical role label sequence $l = \{l_1, l_2,\ldots,l_m\}$  with $l_i \in L$. $x_i$ and $x_{jp}$  denote  $i^\text{th}$ sentence and $p^\text{th}$ token of $j^\text{th}$ sentence, respectively. $m$  and $n$ denote the number of sentences and tokens in the $i^\text{th}$ sentence respectively. $l_i$ is the rhetorical role corresponding to sentence $x_i$ and $L$ denotes set of pre-defined rhetorical role labels. %Further, we also describe the three data augmentation strategies employed. 

\subsection{Hierarchical semi-Markov CRF model}
Our model contains a semi-Markov CRF component \cite{sarawagi2004semi} built on top of a Hierarchical Sequential Labeling Network model \cite{jin2018hierarchical} with the following layers:

\noindent \textbf{Encoding layers:} Following \cite{kalamkar2022corpus}, we encode each sentence with BERT-BASE \cite{kenton2019bert}   to obtain token level representations $z_i = \{z_{i1}, z_{i2},\ldots, z_{in}\}$. These are passed through a Bi-LSTM layer \cite{hochreiter1997long} followed by an attention pooling layer \cite{yang2016hierarchical} to obtain sentence representations $s = \{s_1 , s_2 ,\ldots,s_m \}$.
\begin{equation}
\scalemath{0.9}{
    u_{it} = \tanh(W_w z_{it} + b_w )  ~~\& ~~ 
    \alpha_{it} = \frac{\exp(u_{it}u_w)}{\sum_s \exp(u_{is}u_w)}  ~~\& ~~ 
    s_i = \sum_{t=1}^n \alpha_{it}u_{it}
}
\label{att1}
\end{equation}
where $W_w$, $b_w$, $u_w$ are trainable parameters. \\
\noindent \textbf{Context enrichment layer:} The sentence representations $s$ are passed through a Bi-LSTM to obtain contextualized sentence representations $c=\{c_1 , c_2 , \dots, c_m \}$, which encode contextual information from surrounding sentences.

\noindent \textbf{Classification layer:} A semi-Markov CRF takes the sequence of sentence representations $c$ and segments it into labeled spans $k = \{k_1, ...,k_{|s|} \}$ with $k_j = (a_j,b_j,y_j)$
where $a_j$ and $b_j$ are the starting and ending position of the sentences in the $j^\text{th}$ span, and $y_j$ is the corresponding rhetorical role label of the $j^\text{th}$ span. $|s|$ denotes the total number of spans where $\sum_{l=1}^{|s|}(b_j - a_j +1) = m$.

\noindent We model the conditional probability through a semi-Markov CRF which jointly tackles the span segmentation and label assignment for a span as follows:
\begin{equation}
\scalemath{0.9}{
    p(y|c)=  \frac{1}{Z(c)}\exp( \sum_{j=1}^{|s|} F(k_j, c) + A(y_{j-1},y_j) )
    }\end{equation}
\begin{equation}
\scalemath{0.9}{
~~~\text{where}~~~
    Z(c) =  \sum_{k' \in K} \exp( \sum_{j} F(k'_j, c) + A(y_{j-1},y_j) )
}\end{equation}
where $F(k_j,c)$ is the score assigned for span $k_j$ (i.e., for interval $[a_j, b_j]$ belonging to label $y_j$  based on span input $c$) and $A(y_{j-1},y_j)$ is the transition score of the labels of two adjacent spans. $Z(c)$ denotes the normalization factor computed as the sum over the set of all possible spans $K$ against $c$. The score $F(k_j,c)$ is computed using a learnable weight and bias matrix. 
\begin{equation}
\scalemath{0.9}{
    F(k_j , c) =  W^T. f(k_j , c) + b
}
\end{equation}
where W and b denote trainable parameters and $f(k_j,c)$ represents span representation of $j^\text{th}$ span derived from c. 

To obtain the span representations $f(k_j, c)$, we pass the sentence-level representations $c$ for the sentences in the given span $k_j$ through a BiLSTM layer initially to capture the context of the span. Then we obtain the span representation $f(k_j, c)$ as the concatenation of the first two and final two sentences vectors, and the mean of the sentences in the span. In case of shorter spans, we repeat the same sentence to match the dimension.

We maximize the above defined conditional log-likelihood to estimate the parameters and train the model end-to-end. We perform inference using the  Viterbi decoding algorithm \cite{forney1973viterbi} to obtain the best possible span sequence along with its label assignment. These computations are done in logarithmic space to avoid numerical instability. 
In traditional semi-Markov CRF which are applied to relatively shorter sequences in the previous works, the assumption is that that there exists no transition between the same rhetorical labels.  However, due to the long input data and a larger range of  potential label spans, we relax this assumption as we can deal with a certain maximum  span length due to computational constraints as it involves quadratic complexity.

\subsection{Data Augmentation}
The main goal of Data Augmentation in low resource settings is to increase the diversity of training data which in turn helps the model to generalize better on test data. In this regard,  we implement the following three Data Augmentation techniques as preliminary analysis and leave the exploration of more advanced techniques as a future work. \\
\noindent \textbf{Word deletion} \cite{wei2019eda} is a noise based method that deletes words within a sentence at random. The augmented data differs from the original  without affecting the rhetorical role of the sentence as the rhetorical role of the sentence can be derived from the other words present in the sentence. This helps the model to derive better contextual understanding of the sentence rather than relying on word-level surface features. \\
\noindent In \textbf{back-translation} \cite{lowell2021unsupervised}, we translate the original text at sentence level into other languages and then back to the original language to obtain augmented data. Unlike word level methods, this method does not not directly deal with individual words but rewrites the whole sentence. This makes the model robust to any writing style based spuriously correlated features and learn the semantic information conveyed by the text.  \\
\textbf{Sentence swapping} \cite{dai2020analysis} is based on the notion that a minor change in order of sentences is still readable for humans. We restrict swapping of sentences to those within a single span, which preserves the overall discourse flow of the document. While some discontinuities will be introduced, the text remains content complete and rhetorical roles do not change. This helps the model to learn the discourse flow of the document and makes the model overcome the limitation of having transition between same spans as described in the previous sub-section.

\section{Experiments \& Discussion}
\noindent \textbf{Datasets :} We experiment on two datasets - (i)  BUILDNyAI dataset \cite{kalamkar2022corpus}  consisting of judgement documents from the Indian supreme court, high court and district courts. It consists of publicly available train and validation splits with 184 and 30 documents, respectively, annotated with 12 different rhetorical role labels along with `None'. As  test dataset is not publicly available, we split and use training dataset for both training and validation and test it on the validation partition; (ii) the BVA PTSD dataset  \cite{walker2019automatic} consists of 25 decisions \footnote{The dataset actually contains 75 decisions, out of which only 25 documents have annotation label for every sentence} by the U.S. Board of Veterans’ Appeals (BVA) from appealed disability claims by veterans for service-related post-traumatic stress disorder (PTSD). We use 19 documents for training and validation, and 6 as test. They are annotated with 5 rhetorical roles along with `None'.   \\

\noindent \textbf{Baselines :} We compare our method, \emph{HSLN-spanCRF+DA (data augmentation)} against the following variants : \emph{HSLN-CRF} (normal CRF, no DA), \emph{HSLN-spanCRF} (spanCRF, no DA) and \emph{HSLN-CRF+DA} (normal CRF with DA). \\

\noindent \textbf{Metrics :} We use both span-macro-F1 and span-micro-F1, which is computed based on match of span-by-span labels \footnote{We post-process and merge the same consecutive labels to obtain the span labels.} (i.e., it encompasses both segmentation into exact spans as well their labeling). We also report span-segmentation-F1 which only evaluates on segmentation of spans ignoring the label. We further evaluate at the sentence level using micro-F1 and macro-F1 following previous works \cite{kalamkar2022corpus}. \\

\noindent \textbf{Implementation Details :}
We use the hyperparameters of \cite{kalamkar2022corpus} for the HSLN model. For the semi-Markov CRF, we obtain the the maximum segment length using validation set and set it to 30 and 4 for BUILDNyAI and BVA datasets respectively. We used a batch size of 1 and trained our model end-to-end using Adam \cite{kingma2014adam} optimizer with a learning rate of 1e-5.  For data augmentation, we employed a maximum word deletion rate of 20\%. For back-translation, we used English, German and Spanish as the sequence of languages. We augmented the dataset once using each DA technique and thus models with DA component were trained with four times the size of training dataset.

\begin{table}
\caption{Model performance on BUILDNyAI and BVA datasets}
\scalebox{0.98}{
  \centering
    \begin{tabular}%{|p{3cm}||p{1cm}|p{1.1cm}|p{1cm}||p{1cm}|p{1cm}||p{1cm}|p{1.1cm}|p{1cm}||p{1cm}|p{1cm}|}
    {|l||c|c|c||c|c||c|c|c||c|c|}
  \hline
    &  \multicolumn{5}{c||}{\textbf{BUILDNyAI }} & \multicolumn{5}{c|}{\textbf{BVA PTSD }}\\
  \hline
   &  \multicolumn{3}{c||}{\textbf{Span }}& \multicolumn{2}{c||}{\textbf{Sentence }}  & \multicolumn{3}{c||}{\textbf{Span }} &  \multicolumn{2}{c|}{\textbf{Sentence }}\\
  \hline
   \textbf{Model}  & \textbf{s-mic.} &\textbf{s-mac.} & \textbf{s-seg} & \textbf{mic.} &\textbf{mac.}  & \textbf{s-mic.} &\textbf{s-mac.} & \textbf{s-seg} &\textbf{mic.} &\textbf{mac.} \\ 
 \hline
CRF  & 	0.31& 0.28& 0.33& 0.80 & 0.60 & 0.67 & 0.58 & 0.71 & 0.81 & 0.74  \\
spanCRF & 0.38& 0.35& 0.39 & 0.76 & 0.56 & 0.67 & 0.56 & 0.69 & 0.78 & 0.72  \\
CRF + DA & 0.32	& 0.32 & 0.34 & \textbf{0.82} & \textbf{0.63} &  0.72 & 0.64 & \textbf{0.75} & \textbf{0.85} & \textbf{0.81} \\
spanCRF + DA & \textbf{0.40} & \textbf{0.36} & \textbf{0.41} & 0.81 & 0.58	 & \textbf{0.73} & \textbf{0.65} & \textbf{0.75} & 0.83 & 0.80  \\
 \hline
  \end{tabular}}
  \label{peformance}
\end{table}

\subsubsection{Performance Evaluation :}
Table \ref{peformance} reports the performance of our model and its variants on the two datasets. On BUILDNyAI, we observe that spanCRF performs better compared to a normal CRF in span-level metrics (statistically significant (p $\le$ 0.05) using McNemar Test), with a drop at the sentence-level. With the addition of data augmentation (DA), both CRF and spanCRF performance improves. However, the increase is larger for spanCRF's sentence level metrics (statistically significant (p $\le$ 0.05) using McNemar Test). This can be attributed to spanCRF having to compute the optimal segmentation path over all the possible paths, which requires enough data to learn and generalize better. On the other hand, on the BVA PTSD dataset, spanCRF did not show a significant impromavement  compared to normal CRF. This is because 73.8\% of the spans in BVA dataset (BUILDNyAI: 31\%) have length 1 and the mean span length is 1.85 (BUILDNyAI: 6.81) which does not allow spanCRF to show its potential. However, the trend towards a beneficial effect of data augmentation persists.  \\

\noindent \textbf{Effect of Maximum Span Length :} We create variants of spanCRF by varying the maximum span length. First section in Table \ref{length} shows that increasing the span length improved the performance on span-level metrics with a marginal drop at the sentence-level. We choose 30 as the maximum span length due to the computational resource constraints and our very long judgment documents. \\

\noindent \textbf{Effect of Span representation :} We experiment with various span representations such as \emph{grConv} \cite{kemos2019neural} (Gated Recursive Convolutional Neural Networks),  \emph{simple} \cite{yamada2020sequential} involving concatenation of first and last sentence representation in span. We also create a variant of our proposed span representation by removing the BiLSTM (\emph{ours w/o BiLSTM}). From second section in Table \ref{length}, we observe a performance drop without the BiLSTM layer (both at span- and sentence-level) indicating the importance of capturing context specifically at the span level to obtain good representations. We notice less improvement with \emph{grConv}, which can also be attributed to its high number of parameters for our low data condition. Though \emph{simple} achieves an improvement in span-level metrics, it shows a huge drop in sentence-level performance. \\

\begin{table}
  \caption{First and second section indicates the effect of max span length (w/o DA) and different span feature representations (w/o DA)   on BUILDNyAI  
%{\color{red} TO UPDATE }
}
\centering
   \begin{tabular}
   %{|p{5cm}||p{1.3cm}|p{1.3cm}|p{1.3cm}||p{1.3cm}|p{1.3cm}|}
   {|l||c|c|c||c|c|}
  \hline
   &  \multicolumn{3}{c||}{\textbf{Span }}& \multicolumn{2}{c||}{\textbf{Sentence }}  \\
  \hline
   \textbf{Model}  & \textbf{s-mic.} &\textbf{s-mac.} & \textbf{s-seg} & \textbf{mic.} &\textbf{mac.}   \\ 
 \hline
CRF (len = 1) & 0.31& 0.28& 0.33& \textbf{0.80} & \textbf{0.60} \\
spanCRF (len=5) & 0.33 & 0.30 & 0.34 & 0.68	& 0.45	 \\
spanCRF (len=10)& 0.34& 0.32 & 0.36	& 0.71 & 0.48	 \\
spanCRF (len=20) & 0.36	& 0.33 & 0.37 & 0.73 &	0.52	\\
spanCRF (len=30) &  \textbf{0.38}& \textbf{0.35} & \textbf{0.39} & 0.76 & 0.56 \\
 \hline
CRF (no span) & 0.31& 0.28& 0.33& \textbf{0.80} & \textbf{0.60} \\
Span CRF (ours)  & \textbf{0.38}& \textbf{0.35} & \textbf{0.39} & 0.76 & 0.56 \\
Span CRF (ours w/o BiLSTM) &  0.36 &	0.32 &	0.37 &  0.75 & 0.55 \\
Span CRF (grConv)  & 0.32	& 0.30 & 	0.34 &  0.74 &  0.51 \\
Span CRF (simple)  &  0.34 & 0.33	& 0.36 &  0.72	& 0.52 \\
 \hline
  \end{tabular}
  \label{length}
\end{table}

\noindent \textbf{Ablation on Data Augmentation Strategies :} We observe the effect of each data augmentation strategy in isolation.  From Table \ref{da}, we observe that, in the case of CRF, each of the augmentation strategies boosted performance at the sentence-level by a considerable margin. With all three augmentation strategies combined, CRF witnessed a considerable jump, indicating the complementarity between the strategies. Similarly, we observe an improvement with each data augmentation strategy in case of spanCRF, and the greatest increase when using all three strategies combined.

\begin{table}
\caption{Different data augmentations on CRF and spanCRF on BUILDNyAI}
\scalebox{0.9}{
\centering
    \begin{tabular}
    %{|p{3cm}||p{0.85cm}|p{1.2cm}|p{0.85cm}|p{1.2cm}|p{0.85cm}|p{1.2cm}||p{0.85cm}|p{1.2cm}|p{0.85cm}|p{1.2cm}|}
    {|l||c|c|c|c|c|c||c|c|c|c|}
  \hline
   &  \multicolumn{6}{c||}{\textbf{Span }}& \multicolumn{4}{c|}{\textbf{Sentence }}  \\
  \hline
    &  \multicolumn{2}{c|}{\textbf{s-mic.}} &  \multicolumn{2}{c|}{\textbf{s-mac.}} &  \multicolumn{2}{c||}{\textbf{s-seg}} &  \multicolumn{2}{c|}{\textbf{mic.}} &  \multicolumn{2}{c|}{\textbf{mac.}}   \\ 
 \hline
   \textbf{Model}  & CRF & sp.CRF &CRF & sp.CRF &CRF & sp.CRF &CRF & sp.CRF &CRF & sp.CRF    \\ 
 \hline
No Augmentation & 0.31& 0.38&  0.28& 0.35&  0.33& 0.39 &  0.80 & 0.76 &  0.60  & 0.56 \\

+ Swapping  & \textbf{0.32} & 0.39 & 0.30 & \textbf{0.36} & \textbf{0.34} & 0.40 & \textbf{0.82} & 0.80  & 0.62 & \textbf{0.58}\\
+ Deletion   & \textbf{0.32} & 0.39  & 0.30 &\textbf{0.36} & \textbf{0.34} & 0.40 & 0.81 & 0.78  & 0.61 & \textbf{0.58}\\
+ Back translation   &\textbf{0.32} & \textbf{0.40} & 0.31 & \textbf{0.36} & \textbf{0.34}  & 0.40 & 0.81 & 0.77 & 0.62 & 0.57 \\
+ All three DA  & \textbf{0.32}	& \textbf{0.40}  & \textbf{0.32} &  \textbf{0.36} & \textbf{0.34} & \textbf{0.41} & \textbf{0.82} & \textbf{0.81} & \textbf{0.63} & \textbf{0.58} \\ 
 \hline
  \end{tabular}}
  \label{da}
\end{table}

\section{Conclusion}
Our experiments demonstrate that while semi-Markov CRFs help to boost the predictions at the span level, data augmentation strategies can mitigate data scarcity and improve the performance both at sentence- and span-levels, albeit conditioned on the documents exhibiting patterns of longer passages of the same rhetorical type. While this is typical for legal judgments, it is not universal. In the future, we hence would like to  combine the complimentary sentence- and span-level methods. We would also like to explore different data augmentation strategies to alleviate the bottle neck of limited annotated data and expensive data annotation, especially in these specialized domains.

\bibliographystyle{splncs04}
\bibliography{custom}

\begin{thebibliography}{10}
\providecommand{\url}[1]{\texttt{#1}}
\providecommand{\urlprefix}{URL }
\providecommand{\doi}[1]{https://doi.org/#1}

\bibitem{ahmad2020understanding}
Ahmad, S.R., Harris, D., Sahibzada, I.: Understanding legal documents:
  classification of rhetorical role of sentences using deep learning and
  natural language processing. In: 2020 IEEE 14th International Conference on
  Semantic Computing (ICSC). pp. 464--467. IEEE (2020)

\bibitem{arora2019semi}
Arora, R., Tsai, C.T., Tsereteli, K., Kambadur, P., Yang, Y.: A semi-markov
  structured support vector machine model for high-precision named entity
  recognition. In: Proceedings of the 57th Annual Meeting of the Association
  for Computational Linguistics. pp. 5862--5866 (2019)

\bibitem{bhattacharya2021deeprhole}
Bhattacharya, P., Paul, S., Ghosh, K., Ghosh, S., Wyner, A.: Deeprhole: deep
  learning for rhetorical role labeling of sentences in legal case documents.
  Artificial Intelligence and Law pp. 1--38 (2021)

\bibitem{dai2020analysis}
Dai, X., Adel, H.: An analysis of simple data augmentation for named entity
  recognition. In: Proceedings of the 28th International Conference on
  Computational Linguistics. pp. 3861--3867 (2020)

\bibitem{farzindar2004letsum}
Farzindar, A., Lapalme, G.: Letsum, an automatic legal text summarizing. In:
  Legal knowledge and information systems: JURIX 2004, the seventeenth annual
  conference. vol.~120, p.~11. IOS Press (2004)

\bibitem{feng2021survey}
Feng, S.Y., Gangal, V., Wei, J., Chandar, S., Vosoughi, S., Mitamura, T., Hovy,
  E.: A survey of data augmentation approaches for nlp. In: Findings of the
  Association for Computational Linguistics: ACL-IJCNLP 2021. pp. 968--988
  (2021)

\bibitem{forney1973viterbi}
Forney, G.D.: The viterbi algorithm. Proceedings of the IEEE  \textbf{61}(3),
  268--278 (1973)

\bibitem{ghosh2019identification}
Ghosh, S., Wyner, A.: Identification of rhetorical roles of sentences in indian
  legal judgments. In: Legal Knowledge and Information Systems: JURIX 2019: The
  Thirty-second Annual Conference. vol.~322, p.~3. IOS Press (2019)

\bibitem{hachey2006extractive}
Hachey, B., Grover, C.: Extractive summarisation of legal texts. Artificial
  Intelligence and Law  \textbf{14}(4),  305--345 (2006)

\bibitem{hochreiter1997long}
Hochreiter, S., Schmidhuber, J.: Long short-term memory. Neural computation
  \textbf{9}(8),  1735--1780 (1997)

\bibitem{jin2018hierarchical}
Jin, D., Szolovits, P.: Hierarchical neural networks for sequential sentence
  classification in medical scientific abstracts. In: Proceedings of the 2018
  Conference on Empirical Methods in Natural Language Processing. pp.
  3100--3109 (2018)

\bibitem{kalamkar2022corpus}
Kalamkar, P., Tiwari, A., Agarwal, A., Karn, S.M., Gupta, S., Raghavan, V.,
  Modi, A.: Corpus for automatic structuring of legal documents. In: LREC
  (2022)

\bibitem{kemos2019neural}
Kemos, A., Adel, H.: Neural semi-markov conditional random fields for robust
  character-based part-of-speech tagging. In: Proceedings of NAACL-HLT. pp.
  2736--2743 (2019)

\bibitem{kenton2019bert}
Kenton, J.D.M.W.C., Toutanova, L.K.: Bert: Pre-training of deep bidirectional
  transformers for language understanding. In: Proceedings of NAACL-HLT. pp.
  4171--4186 (2019)

\bibitem{kingma2014adam}
Kingma, D.P., Ba, J.: Adam: {A} method for stochastic optimization. In: Bengio,
  Y., LeCun, Y. (eds.) 3rd International Conference on Learning
  Representations, {ICLR} 2015, San Diego, CA, USA, May 7-9, 2015, Conference
  Track Proceedings (2015)

\bibitem{kong2015segmental}
Kong, L., Dyer, C., Smith, N.A.: Segmental recurrent neural networks. In:
  Bengio, Y., LeCun, Y. (eds.) 4th International Conference on Learning
  Representations, {ICLR} 2016, San Juan, Puerto Rico, May 2-4, 2016,
  Conference Track Proceedings (2016)

\bibitem{liu2016exploring}
Liu, Y., Che, W., Guo, J., Qin, B., Liu, T.: Exploring segment representations
  for neural segmentation models. In: Proceedings of the Twenty-Fifth
  International Joint Conference on Artificial Intelligence. pp. 2880--2886
  (2016)

\bibitem{lowell2021unsupervised}
Lowell, D., Howard, B., Lipton, Z.C., Wallace, B.C.: Unsupervised data
  augmentation with naive augmentation and without unlabeled data. In:
  Proceedings of the 2021 Conference on Empirical Methods in Natural Language
  Processing. pp. 4992--5001 (2021)

\bibitem{lu2016segmental}
Lu, L., Kong, L., Dyer, C., Smith, N.A., Renals, S.: Segmental recurrent neural
  networks for end-to-end speech recognition. Interspeech 2016 pp. 385--389
  (2016)

\bibitem{masada2017chord}
Masada, K., Bunescu, R.C.: Chord recognition in symbolic music using
  semi-markov conditional random fields.

\bibitem{nejadgholi2017semi}
Nejadgholi, I., Bougueng, R., Witherspoon, S.: A semi-supervised training
  method for semantic search of legal facts in canadian immigration cases. In:
  JURIX. pp. 125--134 (2017)

\bibitem{saravanan2008automatic}
Saravanan, M., Ravindran, B., Raman, S.: Automatic identification of rhetorical
  roles using conditional random fields for legal document summarization. In:
  Proceedings of the Third International Joint Conference on Natural Language
  Processing: Volume-I (2008)

\bibitem{sarawagi2004semi}
Sarawagi, S., Cohen, W.W.: Semi-markov conditional random fields for
  information extraction. Advances in neural information processing systems
  \textbf{17} (2004)

\bibitem{savelka2018segmenting}
Savelka, J., Ashley, K.D.: Segmenting us court decisions into functional and
  issue specific parts. In: JURIX. pp. 111--120 (2018)

\bibitem{walker2019automatic}
Walker, V.R., Pillaipakkamnatt, K., Davidson, A.M., Linares, M., Pesce, D.J.:
  Automatic classification of rhetorical roles for sentences: Comparing
  rule-based scripts with machine learning. In: ASAIL@ ICAIL (2019)

\bibitem{wei2019eda}
Wei, J., Zou, K.: Eda: Easy data augmentation techniques for boosting
  performance on text classification tasks. In: Proceedings of the 2019
  Conference on Empirical Methods in Natural Language Processing and the 9th
  International Joint Conference on Natural Language Processing (EMNLP-IJCNLP).
  pp. 6382--6388 (2019)

\bibitem{yamada2019neural}
Yamada, H., Teufel, S., Tokunaga, T.: Neural network based rhetorical status
  classification for japanese judgment documents. In: Legal Knowledge and
  Information Systems, pp. 133--142. IOS Press (2019)

\bibitem{yamada2020sequential}
Yamada, K., Hirao, T., Sasano, R., Takeda, K., Nagata, M.: Sequential span
  classification with neural semi-markov crfs for biomedical abstracts. In:
  Findings of the Association for Computational Linguistics: EMNLP 2020. pp.
  871--877 (2020)

\bibitem{yan2021skipping}
Yan, Y., Cwitkowitz, F., Duan, Z.: Skipping the frame-level: Event-based piano
  transcription with neural semi-crfs. Advances in Neural Information
  Processing Systems  \textbf{34},  20583--20595 (2021)

\bibitem{yang2016hierarchical}
Yang, Z., Yang, D., Dyer, C., He, X., Smola, A., Hovy, E.: Hierarchical
  attention networks for document classification. In: Proceedings of the 2016
  conference of the North American chapter of the association for computational
  linguistics: human language technologies. pp. 1480--1489 (2016)

\bibitem{ye2018hybrid}
Ye, Z., Ling, Z.H.: Hybrid semi-markov crf for neural sequence labeling. In:
  Proceedings of the 56th Annual Meeting of the Association for Computational
  Linguistics (Volume 2: Short Papers). pp. 235--240 (2018)

\bibitem{zhuo2016segment}
Zhuo, J., Cao, Y., Zhu, J., Zhang, B., Nie, Z.: Segment-level sequence modeling
  using gated recursive semi-markov conditional random fields. In: Proceedings
  of the 54th Annual Meeting of the Association for Computational Linguistics
  (Volume 1: Long Papers). pp. 1413--1423 (2016)

\end{thebibliography}

\end{document}